\documentclass[sigconf]{acmart}
\usepackage{multirow} 
\usepackage{multicol} 
\usepackage{array}
\usepackage{graphicx}
\usepackage{subfigure}
\usepackage{rotating} 
\usepackage{caption} 
\usepackage{booktabs} 
\usepackage{amsmath} 
\usepackage{amsfonts} 
\usepackage{colortbl} 
\usepackage{xcolor} 
\AtBeginDocument{%
  }

\setcopyright{acmlicensed}
\copyrightyear{2024}
\acmYear{2024}
\acmDOI{XXXXXXX.XXXXXXX}

\acmConference[KDD '24]{Make sure to enter the correct
  conference title from your rights confirmation emai}{August 25--29,
  2024}{Barcelona, Spain}
\acmISBN{978-1-4503-XXXX-X/18/06}

\begin{document}


\title{High-Discriminative Attribute Feature Learning for Generalized Zero-Shot Learning}



\author{Yu Lei}
\affiliation{%
 \institution{Xidian University}
  \city{Xi'an}
  \country{China}}

\author{Guoshuai Sheng}
\affiliation{%
  \institution{Xidian University}
  \city{Xi'an}
  \country{China}}

\author{Fangfang Li}
\affiliation{%
  \institution{Xidian University}
  \city{Xi'an}
  \country{China}}

\author{Quanxue Gao*}
\affiliation{%
  \institution{Xidian University}
 \city{Xi'an}
  \country{China}}
\email{qxgao@xidian.edu.cn}

\author{Cheng Deng}
\affiliation{%
  \institution{Xidian University}
  \city{Xi'an}
  \country{China}}

\author{Qin Li}
\affiliation{%
  \institution{Shenzhen Institute of Information Technology}
  \city{Shenzhen}
  \country{China}}



\begin{abstract}
  Zero-shot learning(ZSL) aims to recognize new classes without prior exposure to their samples, relying on semantic knowledge from observed classes. However, current attention-based models may overlook the transferability of visual features and the distinctiveness of attribute localization when learning regional features in images. Additionally, they often overlook shared attributes among different objects. Highly discriminative attribute features are crucial for identifying and distinguishing unseen classes. To address these issues, we propose an innovative approach called High-Discriminative Attribute Feature Learning for Generalized Zero-Shot Learning (HDAFL). HDAFL optimizes visual features by learning attribute features to obtain discriminative visual embeddings. Specifically, HDAFL utilizes multiple convolutional kernels to automatically learn discriminative regions highly correlated with attributes in images, eliminating irrelevant interference in image features. Furthermore, we introduce a Transformer-based attribute discrimination encoder to enhance the discriminative capability among attributes. Simultaneously, the method employs contrastive loss to alleviate dataset biases and enhance the transferability of visual features, facilitating better semantic transfer between seen and unseen classes. Experimental results demonstrate the effectiveness of HDAFL across three widely used datasets.
\end{abstract}


 %


\maketitle

\section{Introduction}
Human beings possess a remarkable ability to discern and identify objects even when encountering novel ones. Nevertheless, conventional deep learning models often encounter challenges in recognizing classes not present in their training data, severely constraining the capacity of deep learning to emulate human-like cognition. The concept of Zero-Shot Learning (ZSL) \cite{Lampert2009LearningTD} has emerged. ZSL offers a solution to the recognition of new classes by leveraging the inherent semantic relationships discovered during the learning process.

\begin{figure}[t]
	\centering
	\includegraphics[width=0.4\textwidth]{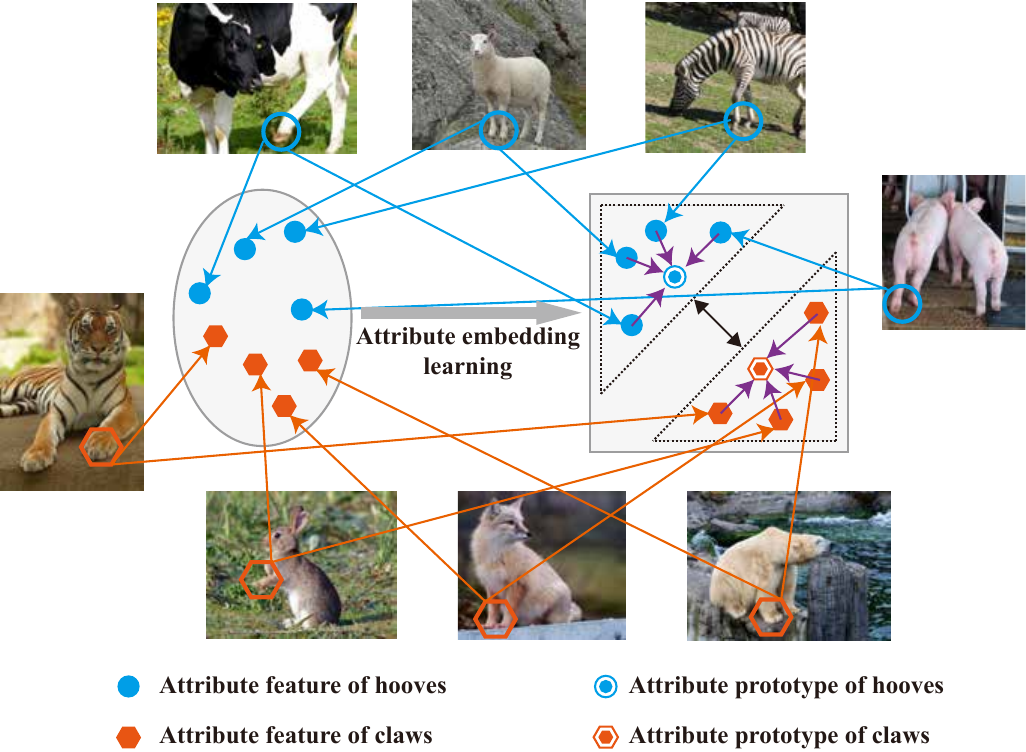}
	\caption{Schematic diagram of attribute embedding space. The previous approach failed to effectively distinguish between different objects with similar attributes in the embedding space(left circle), leading to domain shift issues. In order to address this problem, we propose a method that aims to enhance the embedding space's ability to capture both commonalities and differences among objects (right rectangle), thereby improving discriminability.}\label{fig1}
\end{figure}

In the last decade, substantial advancements have been achieved in the realm of ZSL, which is dedicated to the identification of new classes through the exploitation of intrinsic semantic connections between both seen and unseen classes. In ZSL, the absence of training samples for unseen classes within the test set, coupled with the disjoint nature of label spaces between the training and test sets, presents a unique challenge. In this scenario, where only semantic descriptions of both seen and unseen classes are available, along with training images solely for seen classes \cite{liu2018generalized}, the primary objective of ZSL is to accurately classify test images pertaining to unseen classes.

ZSL methods can be divided into two types based on the label space's different ranges in testing: conventional ZSL (CZSL) and generalized ZSL (GZSL) \cite{chao2016empirical}. CZSL is designed to predict classes that have not been seen before, whereas GZSL has the ability to make predictions for both seen and unseen classes. The semantic descriptions (attributes \cite{lampert2013attribute}) play a crucial role as shared information between these seen and unseen classes, ensuring effective knowledge transfer.

In order to facilitate the transfer of knowledge from seen to unseen classes through visual-semantic interactions, early ZSL methods based on embeddings \cite{song2018transductive}, \cite{akata2015label,li2018discriminative} have aimed to learn embeddings between seen classes and their corresponding class semantic vectors. Subsequently, these methods classify unseen classes by conducting nearest neighbor searches within the embedding space. Nevertheless, these embedding-based methods tend to suffer from overfitting to seen classes when operating in the generalized ZSL (GZSL) scenario, commonly referred to as the bias problem. This issue arises due to embeddings being solely trained using samples from seen classes. To address this problem, a multitude of generative ZSL methods have arisen. They concentrate on synthesizing feature samples for unseen classes by utilizing generative models such as variational autoencoders (VAEs)\cite{verma2018generalized} , generative adversarial networks (GANs)\cite{chen2021free,keshari2020generalized,vyas2020leveraging,xian2018feature,yu2020episode} to augment the available data. Therefore, the generative ZSL methods serve to address the insufficient training data for unseen classes, effectively converting the ZSL problem into a supervised classification task.

Despite their considerable advancements, these methods heavily depend on global visual features, which fall short in capturing the fine-grained attribute information (such as the specific attribute "bill color yellow" of a Red-Legged Kittiwake) essential for effectively representing semantic knowledge [34], [35], [36]. The effectiveness of knowledge transfer from seen to unseen classes is often attributed to discriminative and transferable semantic knowledge within specific regions corresponding to distinct attributes. Consequently, these methods yield suboptimal visual feature representations, leading to a deficiency in visual-semantic interaction during the process of knowledge transfer. Lately, there has been a focus on leveraging semantic vectors within attention-based end-to-end models \cite{huynh2020fine,xie2019attentive,xie2020region,zhu2019semantic,xu2020attribute} to enhance the learning of discriminative part features. Nevertheless, these methods have constraints: they concentrate solely on extracting attribute representations from specific images without adequately considering the shared attribute features across different object categories. Consequently, in the embedding space, representations of the same attribute tend to be dispersed, potentially impeding the model's accuracy in identifying attributes for unseen categories, commonly referred to as the domain shift problem \cite{fu2015transductive}. To overcome these limitations, we propose a novel approach. This method involves adjusting the distribution of identical attribute features for different objects in the embedding space, making it more compact (refer to Figure ~\ref{fig1}). Concurrently, features of different attributes are dispersed further to enhance attribute recognition capabilities. Through these adjustments, we aim to better facilitate the model in transferring semantic knowledge in ZSL, ultimately improving attribute recognition performance for unseen categories.

\begin{figure*}[t]
	\centering
	\includegraphics[width=0.6\textwidth]{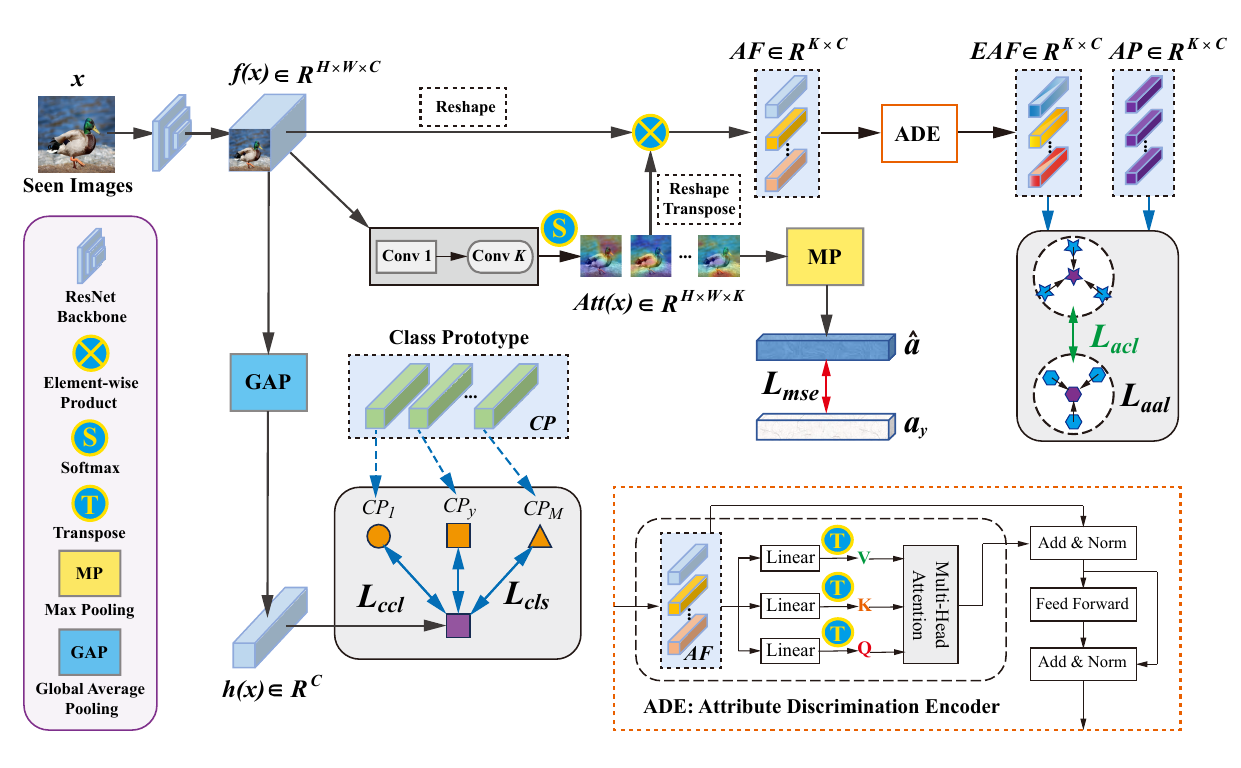}
	\caption{A depiction of the HDAFL, which optimizes visual features by learning attribute features to obtain discriminative visual embeddings.}
\label{fig2}
\end{figure*}

In order to enhance the accuracy of local features in images and mitigate the impact of domain shift in ZSL and GZSL tasks, we propose an embedded-based framework known as the High-Discriminative Attribute Feature Learning model (HDAFL). The overall architecture of the framework is illustrated in Figure ~\ref{fig2}. Specifically, the framework initially utilizes various convolutional kernels to learn local regions corresponding to attributes most relevant to the task. To address potential biases in the semantic space when different objects share similar attributes, HDAFL employs attribute alignment loss and attribute-based contrastive learning loss. These losses aim to align the learned attribute features with the corresponding attribute prototypes, enhancing the representation of similar attributes while reducing confusion between distinct attributes. Furthermore, to augment the discriminative nature of attribute features, we introduce an attribute discrimination encoder based on a transformer structure to obtain refined attribute representations. Additionally, we extract class-level features from images and ensure proximity between features of the same category through class-based contrastive loss, thereby better preserving semantic relationships between categories. Finally, the overall performance of the model is optimized through classification loss and the corresponding class prototypes. It is noteworthy that during the training process, only class prototypes and attribute prototypes are employed as supervisory signals, without relying on human-annotated associative information. This design choice enhances the generalization capabilities of our framework, making it adaptable to various datasets and tasks in the fields of ZSL and GZSL.

The key contributions of HDAFL can be summarized as follows:
\begin{itemize}
	\item The paper proposes an innovative approach named HDAFL. By utilizing attribute features to optimize visual features, it achieves discriminative visual embeddings, addressing potential oversights in current attention models during the learning of image region features.
    \item HDAFL integrates multiple convolutional kernels and the transformer-based attribute discriminative encoder. This combination enables the automatic learning of regions highly correlated with attributes while eliminating irrelevant interference in image features. Consequently, it significantly improves the transferability of visual features.
	\item In order to mitigate dataset bias and enhance the transferability of visual features, HDAFL employs a contrastive loss. This strategy effectively promotes semantic transfer, as demonstrated through successful experiments on three widely used datasets.
\end{itemize}


\section{Related work}
The purpose of ZSL is to transfer the target recognition model from seen to unseen classes by exploiting a semantic space that is shared among them. In this regard, generative ZSL methods, by learning attributes prototypes of class, utilize various generative models such as VAEs \cite{schonfeld2019generalized,wang2018zero}, GAN \cite{su2022distinguishing,xian2018feature,zhu2018generative} , and generative flows \cite{shen2020invertible} to synthesize images or visual features of unseen classes. Despite some success in compensating for the absence of unknown categories, these generative methods turn the ZSL problem into a fully supervised task when additional data is introduced, potentially limiting their applicability in truly ZSL scenarios.

The embedding-based method plays a crucial role in the realm of GZSL, aiming primarily to achieve effective classification of unseen classes through the projection and alignment of visual and semantic information. During the early stages of research, there were efforts to map global visual and semantic features directly to a common embedding space, aiming at category prediction. Nevertheless, these methods often yielded less than satisfactory results. The challenge lay in the difficulty of global visual information capturing subtle yet significant differences between various categories, thereby weakening the model's discriminative power for different classes.

To overcome this issue, attention-based ZSL methods \cite{chen2022transzero,liu2021goal,xie2019attentive,xie2020region,xu2020attribute,zhu2019semantic} have emerged, these models focus more on local regions in the image that are more relevant to attribute descriptions during the learning process. This helps in discovering more fine-grained features and enhances discrimination across different classes. Zhu et al. introduced a model called SGMA \cite{zhu2019semantic}. This model is designed to identify discriminative patches by learning local features, and it shows robust localization performance when utilizing human annotations. Chen et al. \cite{chen2022msdn} presented a novel semantic distillation framework for refining semantic representations in ZSL through mutual teaching and collaborative learning. The Attribute Prototype Network (APN), proposed by Xu et al. \cite{xu2020attribute}, utilizes learnable attribute prototypes to convolve feature maps. In the test phase, calibrated stacking is applied to enhance the performance of GZSL. RGEN \cite{xie2020region} relies on a combination of balanced loss and region-based relation reasoning to ensure a consistent response to both seen and unseen outputs during the training process. Additionally, the GEM model proposed by Liu et al. \cite{liu2021goal} aims to predict human gaze positions, acquire visual attention regions, and identify attribute descriptions of new objects. However, their method relies solely on unidirectional attention, thereby restricting a deeper understanding of semantic alignment between visual and attribute features, lacking any additional sequential learning. Consequently, we emphasize the importance of thoroughly exploring the intrinsic semantic representations of visual and attribute features in ZSL. In conclusion, our research focuses on exploring common features among objects with noticeable visual differences but possessing the same attributes, aiming to improve the performance of ZSL.

\section{Method}
This section begins with defining notation and problem settings, providing a basis for ongoing discussion. We then delve into our proposed HDAFL, examining the specifics of each module to ensure a complete understanding.

\subsection{Notation and Problem Settings}
The goal of ZSL is to recognize new classes through the transfer of semantic knowledge from seen classes (${Y^S}$) to unseen classes (${Y^U}$). The image spaces associated with seen and unseen classes are represented as $X = {X^S} \cup {X^U}$. The ZSL model is trained using the seen classes, the training set is denoted as $S = \{ (x,y,{a_y})|x \in {X^S},y \in {Y^S},{a_y} \in {\phi ^S}\}$, where $x$ represents an image in ${X^S}$, $y$ stands for its class label acquired during the training process, and ${a_y} \in {^K}$ represents the semantic embedding at the class level, annotated with $K$ distinct semantic attributes. Regarding the unseen testing set $U = \{ ({x^u},u,\varphi (u))|{x^u} \in {X^U},u \in {Y^U},{a_u} \in {\phi ^U}\}$, where $u$ stands for unseen class labels. The sets of seen and unseen classes do not overlap, denoted as  ${{Y}^{S}} \cap {{Y}^U} = \emptyset $. The transfer of information between seen and unseen classes is achieved through the use of $\phi  = {\phi ^S} \cup {\phi ^U}$. In traditional ZSL, the objective involves predicting the labels assigned to images from classes that have not been seen before, represented as ${{X}^U} \to {{Y}^U}$. However, when faced with a more realistic and complex GZSL setting, the objective becomes predicting images from classes that are both seen and unseen, meaning ${X} \to {{Y}^U} \cup {{Y}^{S}}$.

\subsection{Overview}
Utilizing the ResNet101 backbone, we obtain feature maps denoted as $f(x) \in {R^{H \times W \times C}}$ from the input image $x$, with $H$, $W$, and $C$ corresponding to the height, width, and channel of the features. After that, a global average pooling operation is employed over both the height ($H$) and width ($W$), yielding a global discriminative feature $h(x) \in {R^C}$, with its corresponding label denoted as $y$. After going through the class semantics encoder, the class semantics vector ${a_y}$ turns into the class prototype ${c{p_y}}$. Unlike the previous method described in \cite{xu2020attribute}, where the computation of class logits involves using dot product on projected visual features and each class embedding, we adopt a different strategy based on \cite{he2016deep}. Our classification loss is determined by cross-entropy and utilizes cosine similarity, offering a reduction in intra-class variance and, consequently, improved generalization in models.
Following this, the classification score function is established as:
\begin{equation}\label{eq1}
	{p_x} = \frac{{\exp (\alpha  \cdot \cos (h(x),c{p_y}))}}{{\sum\limits_{i = 1}^M {\exp } (\alpha  \cdot \cos (h(x),c{p_i}))}},
\end{equation}
where $\alpha$ represents the scaling factor, $M$ denotes the total number of seen and unseen class. Our classification loss is formulated for $B$ images in a batch as follows:
\begin{equation}\label{eq2}
	{L_{{\rm{cls}}}} =  - \frac{1}{B}\sum\limits_{i = 1}^B {\log {p_i}},
\end{equation}


\subsection{Attribute Embedding Learning}

\textbf{Attribute feature extraction.} To obtain attribute features, we first employ $K$ convolution kernels of size $1 \times 1 \times C$ and a softmax operation to map the visual representation $f(x)$ to the attribute feature map $Att(x) \in {R^{H \times W \times K}}$. By performing this operation, we effectively capture and identify the semantic characteristics of each attribute, aligning them with corresponding areas on the attention maps. Subsequently, we proceed with attribute embedding, as depicted in Figure ~\ref{fig2}. It takes the $f(x)$ and the $Att(x)$ as inputs and obtains the attribute feature $AF \in {R^{K \times C}}$ as the output. In the specific implementation, we reshape and transpose $Att(x)$ to have dimensions $K \times HW$, and similarly reshape $f(x)$ to attain a dimension of $HW \times C$. Following this, the attribute features $AF = \{ a{f_i}\} _{i = 1}^K$ are obtained by multiplying the matrices of $f(x)$ and $Att(x)$.

\textbf{Attribute-base feature selection.} To facilitate attribute embedding learning, we utilize attribute features $AF$ to distinguish between different attributes. However, not all channel features are equally important. Before embarking on attribute embedding learning, it is necessary to incorporate feature selection for channels. To address this issue, we introduce an attribute discrimination encoder designed to enhance the features of important channels, as illustrated in the bottom right corner of Figure ~\ref{fig2}. This encoder can learn to allocate different weights for each attribute, allowing it to more effectively focus on channels associated with each attribute. This contributes to enhancing discrimination between different attributes, as the model can selectively emphasize or attenuate information related to each attribute. The attribute feature $AF$ undergoes processing within an encoding network based on the transformer structure, enabling the effective differentiation of information represented by different channels. As a result, enhanced attribute features, denoted as $EAF \in {R^{K \times C}}$, are obtained. Specifically, we first split $AF$ into $h$ blocks along the channel dimension, obtaining $AF = \left\| {_{i = 1}^h} \right.A{F_i}$ (where $A{F_i} \in {R^{K \times \frac{C}{h}}}$), the attribute discrimination encoder based on channel attention is represented as follows:
\begin{equation}\label{eq31}
	Q_i = (AF_i)W^Q_i, K_i = (AF_i)W^K_i, V_i = (AF_i)W^V_i
\end{equation}
where $W^Q_i \in {R^{\frac{C}{h} \times \frac{C}{h}}}$, $W^K_i \in {R^{\frac{C}{h} \times \frac{C}{h}}}$, $W^V_i \in {R^{\frac{C}{h} \times \frac{C}{h}}}$ represent the weight parameters learned by the linear layer. Then, we further transpose $Q_i$, $K_i$, and $V_i$ to obtain ${Q_i}^T$, ${K_i}^T$, and ${V_i}^T$. We can obtain the following equation.
\begin{equation}
	\begin{gathered}
\text{head}_i = \text{softmax}\left(\frac{{Q_i}^TK_i}{\sqrt{K}}\right)V_i^T \\
MH(Q, K, V) = \left\|_{i=1}^n\right.(\text{head}_i^T)W^o \\
\widehat{AF} = AF + MH(Q, K, V)
	\end{gathered}
\end{equation}
where $hea{d_i}^T$ represent the transpositions of $hea{d_i} \in {R^{\frac{C}{h} \times K}}$;  $||$ stands for the concatenation operation. $W^o \in {R^{C \times C}}$ represents the weight parameters learned after concatenating all the heads, and ${\sqrt{K}}$ represents the scaling factor.

Following that, the $\widehat{AF} \in {R^{K \times C}}$ undergoes a feed-forward layer for non-linear transformation of its features, enhancing the model's ability to produce more expressive results. The ultimate output of the network is enhanced attribute features $EAF \in {R^{K \times C}}$.

\textbf{Attribute Features Alignment.} For each input image $x$, we obtain its attribute features $AF$. However, the same attribute may exhibit different forms in different images, adding complexity to addressing the domain shift problem. To tackle this issue, our objective is to capture the consistency in attribute features shared among diverse objects. More precisely, we permit the distribution of attribute features around their respective attribute prototypes $AP \in {R^{K \times C}}$, ensuring that the $j$-th attribute's feature, $a{f_j}$, closely aligns with its attribute prototype $a{p_j}$ within the embedding space. This strategy contributes to sustaining attribute feature uniformity throughout the learning process. When dealing with a batch of images, we initially filter out attribute features absent in certain images, yielding a refined set of eligible attribute features $AF' = \{ a{f_j}'\} _{j = 1}^{K'}$. This step ensures that the model focuses on the attributes each image possesses during the learning process, enhancing the modeling of internal image attributes. Finally, to effectively align attribute features with their corresponding attribute prototypes, our attribute alignment loss is as follows:
\begin{equation}\label{eq3}
	L_{aal} = \sum\limits_{j = 1}^{K'} {{\rm{Relu}}(} \cos (a{f_j}',a{p_j}) - 0.5{\min _{{j^\prime } \ne j}}\cos (a{f_j}',a{p_{{j^\prime }}})),
\end{equation}

\textbf{Attribute-based Contrastive Learning.} In various categories, objects with the same attributes exhibit different visual representations. The heterogeneity in their embedding space gives rise to the challenge of domain shift. One approach to solving this problem is to make visual features with the same attributes closer to each other, while visual features with different attributes are kept farther apart, in order to enhance the discriminability of attributes in the embedding space. Contrastive learning is a way to achieve this goal, but general supervised contrastive learning \cite{khosla2020supervised} has issues such as increased computational time complexity due to the retention of simple samples, as well as learning boundaries between attributes that are relatively ambiguous. we introduce a novel loss function that dynamically selects hard samples during model optimization, prioritizing challenging samples in the learning process. This helps the model learn highly discriminative features in the attribute embedding space, thereby effectively capturing distinctions between different categories. As shown in Figure ~\ref{fig2}, attribute-based contrastive learning aggregates visual features of the same attribute of different objects, which is beneficial for the model to recognize coarse-grained images. To reduce the algorithmic complexity arising from redundant features, we perform attribute filtering and optimization, resulting in features denoted as $AF' = \{ a{f_j}'\} _{j = 1}^{K'}$. In addition, the model employs a hard sample selection strategy that relies on ranking through cosine similarity, indicating that when selecting hard samples, the model tends to focus on samples that are highly similar but challenging to distinguish in the attribute space. Specifically, for each $a{f_j}'$ in ${AF}'$, the cosine similarity with other attribute features possessing the same attributes (i.e., positive samples) is computed and ranked in descending order. Simple positive samples are removed up to the top $\mu$ ($0 \le \mu  < 1$) to obtain hard positive samples $\{ af_{ju}^ + \} _{u = 1}^U$. Similarly, for attribute features with different attributes from $a{f_j}'$ (i.e., negative samples), cosine similarity is computed and ranked in ascending order, removing simple negative samples up to the top $\varepsilon$ ($0 \le \varepsilon  < 1$) to obtain hard negative samples $\{ af_{jv}^ - \} _{v = 1}^V$. Finally, the similarity between $a{f_j}'$ and its hard positive and hard negative samples is calculated($S(a{f_j}',af_{ju}^ + ) = \exp (\cos (a{f_j}',af_{ju}^ + )/\tau )$ and $S(a{f_j}',af_{jv}^ - ) = \exp (\cos (a{f_j}',af_{jv}^ - )/\tau )$). Throughout this process, by filtering positive and negative samples, the algorithm aims to explore and focus on more challenging samples to improve model performance. The temperature parameter $\tau$ is used to regulate the attention given to hard negative samples. In our experiments, we let $\tau  = 0.3$. Inspired by \cite{khosla2020supervised}, our attribute-based contrastive loss for every $a{f_j}'$ in ${AF}'$ is expressed as:
\begin{equation}\label{eq4}
	{p_j} =  - \frac{1}{U}\sum\limits_{u = 1}^U {\log \frac{{S(a{f_j}',af_{ju}^ + )}}{{\sum\limits_{u = 1}^U {S(a{f_j}',af_{ju}^ + )}  + \sum\limits_{v = 1}^V {S(a{f_j}',af_{jv}^ - )} }}},
\end{equation}
where $U$ and $V$ are the number of positives and negatives of $a{f_j}'$, respectively. Our attribute-based contrastive loss for ${AF}'$ is expressed as:
\begin{equation}\label{eq5}
	L_{acl} = \frac{1}{K'}\sum\limits_{j = 1}^{K'} {{p_j}},
\end{equation}

\textbf{Class-base Contrastive Learning.} In order to enhance the discriminative ability of the attribute embedding space across different classes, we introduce a class-based contrastive learning approach. In this method, we aim to improve the model's ability to recognize fine-grained images by bringing closer together the attribute representations of images belonging to the same class. To achieve this, illustrated in Figure ~\ref{fig2}, we extract the global discriminative feature $h(x) \in {R^{C}}$ from the input image $x$. For a batch of images, we identify the eligible set of global discriminative feature as $\{ h{(x)_i}\} _{i = 1}^B$, with $B$ representing the batch size. Similar to the previously mentioned approach, we define the similarity between $h{(x)_i}$ and $h{(x)_p}$ as $S(h{(x)_i},h{(x)_p}) = \exp (\cos (h{(x)_i},h{(x)_p})/\tau )$. Subsequently, the class-based contrastive loss is formulated as:
\begin{equation}\label{eq6}
	L_{ccl} = \frac{1}{B}\sum\limits_{i = 1}^B {\frac{{ - 1}}{{\left| {P(i)} \right|}}\sum\limits_{p \in P(i)} {\log \frac{{S(h{(x)_i},h{(x)_p})}}{{\sum\limits_{a = 1,a \ne i}^B {S(h{(x)_i},h{(x)_a})} }}} },
\end{equation}
where $P(i) \equiv \{ p \in \{ 1,...,B\} :{y_p} = {y_i},p \ne i\}$ represents the set of indices for all positives, and $\left| {P(i)} \right|$ denotes its cardinality. We set the temperature parameter $\tau  = 0.1$ in our experiments.

\textbf{Attribute Localization.} To further enhance the ZSL model's ability to localize, we apply global max pooling operation on $Att(x)$ over the height ($H$) and width ($W$) dimensions. This operation results in the attributes response value $\widehat a$. Following \cite{xu2020attribute}, the optimization of $\widehat a$ is performed through Mean Square Error (MSE), supervised by the ground truth attributes ${a_y}$.

\begin{equation}\label{eq7}
	{L_{mse}} = \left\| {\widehat a - {a_y}} \right\|_2^2,
\end{equation}
where $y$ represents the label corresponding to the image $x$. The minimization of MSE loss plays a crucial role in improving the accuracy of discriminative attribute localization.

\subsection{Optimization}
In our model, the comprehensive loss function is formulated as:

\begin{equation}\label{eq8}
	L = {L_{cls}} + {\lambda _1}{L_{mse}} + {\lambda _2}L_{aal} + {\lambda _3}L_{acl} + {\lambda _4}L_{ccl}
\end{equation}
where the hyperparameters ${\lambda _1}$, ${\lambda 2}$, ${\lambda _3}$ and ${\lambda _4}$ are associated with the mse loss, attribute alignment loss, attribute-based contrastive loss, and class-based contrastive loss, in that order. In the context of GZSL tasks, concurrently training both class representations and attribute embeddings is a pivotal undertaking that has the potential to boost the model's performance.

\subsection{Zero-Shot Recognition}
In the context of ZSL prediction, when provided with a test image $x$ belonging to unseen classes, the task involves predicting the corresponding global discriminative feature $h(x)$ and assessing its compatibility score with class prototypes. The label associated with the highest compatibility score is considered the predicted class label.
\begin{equation}\label{eq9}
	\hat y = \mathop {\mathop {\arg \max \alpha }\limits_{\tilde y \in {{Y}^u}}  \cdot \cos (h(x),c{p_{\tilde y}})}\limits_{},
\end{equation}

In the GZSL scenario, the test images include classes from both seen and unseen. The model, being trained solely on seen classes, will inevitably display bias towards those seen classes\cite{liu2021task} in its predictions. In order to tackle this problem, calibrated stacking (CS) \cite{chao2016empirical} is applied to lower the scores of the seen classes with a calibration factor $\gamma$. The GZSL classifier is formulated as follows:
\begin{equation}\label{eq10}
	\hat y = \mathop {\mathop {\arg \max }\limits_{\tilde y \in {{Y}^u} \cup {{Y}^s}} \alpha  \cdot \cos (h(x),c{p_{\tilde y}})}\limits_{}  - \gamma {\mathbb{I}\left[ {\tilde y \in {{Y}^s}} \right]},
\end{equation}
where $\alpha$ stands for the same parameter as defined in Equation ~\ref{eq1}. $I$ represents an indicator function, yielding 1 when ${\tilde y \in {{Y}^s}}$, and 0 otherwise.

\section{Experiments}

\subsection{Datasets}
Our framework underwent assessment using three prevalent ZSL benchmark datasets, encompassing both fine-grained datasets (like CUB \cite{welinder2010caltech} and SUN \cite{patterson2012sun}) and a coarse-grained dataset (such as AWA2 \cite{xian2018zero}). AWA2 is a dataset that characterizes various animal species. It comprises 37,322 images categorized into 40 seen classes and 10 unseen classes, each class being defined by 85 attributes. CUB is comprised of 11,788 images of birds distributed among 200 classes, with each class defined by 312 attributes. Among these classes, 150 are categorized as seen while 50 are categorized as unseen. SUN, a dataset comprising 14,340 images across 717 scene classes, further classified into 645 seen classes and 72 unseen classes. These classes can each be defined by a detailed 102-dimensional semantic vector, providing comprehensive descriptors for each category. Table I offers more comprehensive statistics for these three datasets. Additionally, our evaluation employs the Proposed Split (PS) \cite{xian2018zero} as it enforces stricter criteria, ensuring no class overlaps with ImageNet classes for a more precise assessment.

\begin{table}[h]
  \centering
  \caption{Statistics for Datasets CUB, SUN, and AWA2, Taking into Account the Granularity, Attributes, Classes and Data Split.}
  \begin{tabular}{ccccc}
    \toprule
    Dataset & Granularity & Attribute & Seen/Unseen & Train/Test \\
    \midrule
    CUB & fine &312 & 150/50 & 7057/4731 \\
    SUN & fine &102 & 645/72 & 10320/4020 \\
    AWA2 & coarse & 85 & 40/10 & 23527/13795 \\
    \bottomrule
  \end{tabular}
  \label{table1}
\end{table}

\subsection{Metrics}

In traditional ZSL, the evaluation of ZSL performance relies on the average per-class Top-1 accuracy (ACC), where the test set exclusively comprises classes that are unseen during training. In the context of GZSL, the test set comprises both seen and unseen images. Consequently, the evaluation involves computing the T1 accuracy separately for test samples belonging to the seen classes (represented as $S$) and the unseen classes (represented as $U$). The assessment of GZSL performance utilizes their harmonic mean, formulated as $H = (2 \times S \times U)/(S + U)$ \cite{xian2018zero}.

\subsection{Implementation Details}
The HDAFL model we propose is capable of end-to-end training. The backbone of our HDAFL model utilizes a ResNet101 \cite{he2016deep} that has been pre-trained on ImageNet-1K \cite{deng2009imagenet}. This pre-training is specifically employed to extract the visual feature map $f(x) \in {R^{H \times W \times C}}$, where $H$ and $W$ represent feature maps’s height and width, and C denotes the number of channels. It's important to note that this extraction is performed without any fine-tuning. The class semantics encoder and attribute semantics encoder in the network are composed of multilayer perceptrons (MLPs), each consisting of two hidden layers, with each layer having a size of $1024$, and an output layer of size $2048$, serving as the subsequent embedding functions. Throughout the training, parameters are updated using an SGD \cite{bottou2010large} optimizer with momentum. The learning rate is established at ${10^{ - 3}}$, the momentum value is set to $0.9$, and weight decay is specified as ${10^{ - 5}}$. The model is trained for $15$ epochs on all three datasets. The calibration factor $\gamma$ is configured at $1.0$ for AWA2, while it stands at $0.7$ for CUB and SUN datasets. The temperature parameter $\tau$, applied to both attribute-based and class-based contrastive loss, is set to $0.3$ and $0.1$. For the selection of challenging samples, $\mu$ and $\varepsilon$ are uniformly set to $0.32$ and $0.42$, respectively, across all datasets. The scale factor $\alpha$ maintains a consistent value of $25$ across these datasets. Our experiments adopt an episode-based training approach, wherein we sample $M$ categories and $N$ images per category within each minibatch. All experiments are conducted using PyTorch on NVIDIA GeForce GTX2080 Ti GPUs.

\subsection{Comparison with State of the Arts}
In order to demonstrate the effectiveness of our proposed approach, we opted to compare it with the most advanced ZSL methods. These methods encompass both generative approaches, such as f-CLSWGAN \cite{xian2018feature}, f-VAEGAN-D2 \cite{xian2019f}, LisGAN \cite{li2019leveraging}, TF-VAEGAN \cite{narayan2020latent}, HSVA \cite{chen2021hsva}, ICCE \cite{kong2022compactness}, as well as embedding-based methods, namely , TCN \cite{jiang2019transferable}, DAZLE \cite{huynh2020fine}, RGEN \cite{xie2020region}, APN \cite{xu2020attribute}, DCEN \cite{wang2021task}, GEM-ZSL \cite{liu2021goal}, CE-GZSL \cite{han2021contrastive}, MSDN \cite{chen2022msdn}, TransZero \cite{chen2022transzero} and HAS \cite{chen2023zero}.

\textbf{Performance on CZSL.} Table \ref{table1} displays a comparative analysis of performance with contemporary ZSL methods across three benchmark datasets. In the context of CZSL, our HDAFL demonstrates superior accuracy, achieving the highest performance across three datasets: $77.9\%$ on CUB, $67.6\%$ on SUN, and $74.6\%$ on AWA2. The noticeable enhancement in accuracy on SUN can be attributed to our method's improved ability to effectively extract commonalities among features belonging to the same attribute. This enables accurate recognition of attributes in previously unseen classes. Our method demonstrates effective fine-grained recognition, as evidenced by its strong performance on CUB. This achievement is a result of our model's ability to develop a distinct attribute embedding space through the process of learning.

\textbf{Performance on GZSL.} The results presented in Table \ref{table1} demonstrate the performance of different methods under the GZSL conditions. Notably, Our HDAFL exhibits superior performance, achieving the highest scores of $74.3\%$, $44.9\%$, and $75.5\%$ on CUB, SUN, and AWA2 for the metrics denoted as $H$. A significant progress is evident in CUB, with a remarkable improvement of $2.5\%$. Notably, our method stands out with the highest top-1 accuracies for unseen classes in CUB datasets when compared to alternative approaches. This provides evidence of our method's ability to successfully tackle the domain shift problem and exhibit greater discriminative power in tasks related to object classification.

\begin{table*}[htbp]
  \centering
  \caption{Results (\%) from the most advanced CZSL and GZSL techniques on CUB, SUN, and AWA2 datasets are showcased.  The top and second-best achievements are highlighted in \textcolor{red}{red} and \textcolor{blue}{blue}, respectively. The absence of results is indicated by the symbol ”–”. The methods are divided into generative-based ZSLs(GEN) and  embedding-based ZSLs(EMB) methods.}
      \begin{tabular}{c|r|c|ccc|c|ccc|c|cccr}			
    \toprule
    \multirow{3}[6]{*}{} & \multicolumn{1}{>{\raggedright}p{10em}|}{\multirow{3}[6]{*}{Methods}} & \multicolumn{4}{c|}{CUB} & \multicolumn{4}{c|}{SUN} & \multicolumn{4}{c}{AWA2} \\
    \cmidrule{3-15}          &       & \multicolumn{1}{c|}{ZSL} & \multicolumn{3}{c|}{GZSL} & \multicolumn{1}{c|}{ZSL} & \multicolumn{3}{c|}{GZSL} & \multicolumn{1}{c|}{ZSL} & \multicolumn{3}{c|}{GZSL} &  \\													
\cmidrule{3-14}          &       & \multicolumn{1}{c|}{\textbf ACC} & \multicolumn{1}{c}{\textbf U} & \multicolumn{1}{c}{\textbf S} & \multicolumn{1}{c|}{\textbf H} & \multicolumn{1}{c|}{\textbf ACC} & \multicolumn{1}{c}{\textbf U} & \multicolumn{1}{c}{\textbf S} & \multicolumn{1}{c|}{\textbf H} & \multicolumn{1}{c|}{\textbf ACC} & \multicolumn{1}{c}{\textbf U} & \multicolumn{1}{c}{\textbf S} & \multicolumn{1}{c}{\textbf H} \\
    \midrule			

        \multicolumn{1}{c|}{\multirow{2}[10]{*}{\begin{turn}{-90}GEN\end{turn}}} 	
     & f-CLSWGAN(CVPR'18) \cite{xian2018feature} & 57.3  & 43.7  & 57.7  & 49.7  & 60.8  & 42.6  & 36.6  & 39.4  & 68.2  & 57.9  & 61.4  & 59.6 \\
     & f-VAEGAN-D2(CVPR’19) \cite{xian2019f} & 61.0   & 48.4  & 60.1  & 53.6  & 64.7  & 45.1  & 38.0    & 41.3  & 71.1  & 57.6  & 70.6  & 63.5 \\
	& LisGAN(CVPR’19) \cite{li2019leveraging}& 58.8  & 46.5  & 57.9  & 51.6  & 61.7  & 42.9  & 37.8  & 40.2  & -     & -     & -     & - \\
		& TF-VAEGAN(ECCV’20) \cite{narayan2020latent}& 64.9  & 52.8  & 64.7  & 58.1  & \textcolor{blue}{66.0}    & 45.6  & \textcolor{blue}{40.7}  & 43.0    & 72.2  & 59.8  & 75.1  & 66.6 \\
		& HSVA(NeurIPS’21) \cite{chen2021hsva} & -     & 52.7  & 58.3  & 55.3  & -     & 48.6  & 39.0    & \textcolor{blue}{43.3}  & -     & 56.7  & 79.8  & 66.3 \\
		& ICCE(CVPR’22) \cite{kong2022compactness} & -     & 67.3  & 65.5  & 66.4  & -     & -     & -     & -     & -     & 65.3  & 82.3  & 72.8 \\
    \midrule
        \multicolumn{1}{c|}{\multirow{3}[20]{*}{\begin{turn}{-90}EBM\end{turn}}} 						
 & TCN(ICCV’19) \cite{jiang2019transferable}  & 59.5  & 52.6  & 52.0    & 52.3  & 61.5  & 31.2  & 37.3  & 34.0    & 71.2  & 61.2  & 65.8  & 63.4 \\
		& DAZLE(CVPR’20) \cite{huynh2020fine} & 66.0    & 56.7  & 59.6  & 58.1  & 59.4  & \textcolor{blue}{52.3}  & 24.3  & 33.2  & 67.9  & 60.3  & 75.7  & 67.1 \\
		& RGEN(ECCV’20) \cite{xie2020region} & 76.1  & 60.0    & 73.5  & 66.1  & 63.8  & 44.0    & 31.7  & 36.8  & \textcolor{blue}{73.6}  & \textcolor{red}{67.1}  & 76.5  & 71.5 \\
		& APN(NeurIPS’20) \cite{xu2020attribute}  & 72.0    & 65.3  & 69.3  & 67.2  & 61.6  & 41.9  & 34.0    & 37.6  & 68.4  & 57.1  & 72.4  & 63.9 \\
		& DCEN(AAAI’21) \cite{wang2021task} & -     & 63.8  & \textcolor{blue}{78.4}  & 70.4 & -     & 43.7  & 39.8  & 41.7  & -     & 62.4  & 81.7  & 70.8 \\
 & GEM-ZSL(CVPR’21) \cite{liu2021goal} & 77.8     & 64.8  & 77.1  & 70.4  & 62.8     & 38.1  & 35.7    & 36.9  & 67.3     & 64.8  & 77.5  & 70.6 \\
& CE-GZSL(CVPR’21) \cite{han2021contrastive} & \textcolor{blue}{77.5}  & 63.9  & 66.8  & 65.3  & 63.3  & 48.8  & 38.6  & 43.1  & 70.4  & 63.1    & 78.6  & 70.0 \\
& MSDN(CVPR’22) \cite{chen2022msdn} & 76.1  & 68.7  & 67.5  & 68.1  & 65.8  & 52.2  & 34.2  & 41.3  & 70.1  & 62.0    & 74.5  & 67.7 \\
  & TransZero(AAAI’22) \cite{chen2022transzero} & 76.8  & 69.3  & 68.3  & 68.8  & 65.6  & \textcolor{red}{52.6}  & 33.4  & 40.8  & 70.1  & 61.3    & 82.3  & 70.2 \\
        & HAS(ACMM'23) \cite{chen2023zero} &76.5  & \textcolor{blue}{69.6}  & 74.1  & \textcolor{blue}{71.8}  & 63.2  & 42.8  & 38.9  & 40.8  & 71.4  & 63.1    & \textcolor{blue}{87.3}  & \textcolor{blue}{73.3} \\
 & \textbf{HDAFL(Ours)} & \textcolor{red}{77.9}    & \textcolor{red}{70.1} & \textcolor{red}{78.9} & \textcolor{red}{74.3} & \textcolor{red}{67.6}    & 48.3 & \textcolor{red}{41.9} & \textcolor{red}{44.9} & \textcolor{red}{74.6}    & \textcolor{blue}{65.5} & \textcolor{red}{88.9} & \textcolor{red}{75.5} \\
    \bottomrule
    \end{tabular}%
  \label{tab:addlabel}%
\end{table*}%

\subsection{Ablation Studies}
\textbf{Component Analysis.} In order to analyze the contribution of each introduced loss, we conduct ablation studies on the AWA2, CUB, and SUN datasets to assess the effectiveness of HDAFL. In Table \ref{table2}, we categorize training strategies into four combinations. The baseline model utilizes ${L_{cls}}$ and ${L_{mse}}$ for HDAFL training, gradually incorporating attribute alignment loss($L_{aal}$), attribute-based contrastive learning loss($L_{acl}$), and class-based contrastive loss($L_{ccl}$). The incremental addition of these components contributes to the improvement of model performance. Results indicate that HDAFL performs optimally when combining all components. In the ZSL setting, HDAFL achieves $ACC$ improvements of $1.2\%$, $1.2\%$, and $2.4\%$ on CUB, SUN, and AWA2 datasets, respectively. In the GZSL setting, $H$ values increase by $2.5\%$ (CUB), $2.4\%$ (SUN), and $0.9\%$ (AWA2). The second row illustrates that adding $L_{aal}$ on top of the baseline model enhances model performance, particularly on the CUB dataset. This is attributed to $L_{aal}$ ability to learn clear attribute prototypes and align attribute features with prototypes. Furthermore, adding $L_{acl}$ on top of the second row helps compact features of the same attribute and establishes clear boundaries between features of different attributes. For the $H$ metric, there is a $1.5\%$ and $1.1\%$ improvement on the CUB and AWA2 dataset, indicating the crucial role of discriminative representations of semantic attributes across different categories and robust representations within the same category for model performance enhancement. Finally, with the addition of $L_{ccl}$, HDAFL achieves improvements of $0.7\%$, $1.3\%$, and $1.4\%$ on CUB, SUN and AWA2 datasets for the $H$ metric, respectively. These components collectively constitute the HDAFL model, significantly enhancing model performance in both ZSL and GZSL settings.

\begin{table}[htbp]
  \centering
  \caption{Analysis of the impact of different components of HDAFL on three datasets in the context of ZSL and GZSL, emphasizing optimal results displayed in bold.}
    \begin{tabular}{l|cc|cc|cc}
    \toprule
    \multirow{2}[4]{*}{Methods} & \multicolumn{2}{c|}{CUB} & \multicolumn{2}{c|}{SUN} & \multicolumn{2}{c}{AWA2} \\
\cmidrule{2-7}          & \textbf{ACC}   & \textbf{H}     & \textbf{ACC}   & \textbf{H}     & \textbf{ACC}   & \textbf{H} \\
    \midrule
    Baseline & 76.7  & 71.8  & 66.4  & 42.5  & 72.2  & 74.6 \\
    $+ {L_{aal}}$    & 76.8  & 72.1  & 66.4  & 42.5  & 72.4  & 74.1 \\
    $+L_{acl}$ & 76.8  & 73.6  & 66.9  & 43.6  & 73.0    & 74.1 \\
    $+L_{ccl}$ (our) & \textbf{77.9} & \textbf{74.3} & \textbf{67.6} & \textbf{44.9} & \textbf{74.6} & \textbf{75.5} \\
    \bottomrule
    \end{tabular}%
  \label{table2}%
\end{table}%

\textbf{Analysis of Training Methods.} In our experiments, we employe an episode-based training method with the aim of enhancing the model's generalization ability. Specifically, for each mini-batch, we randomly select $M$ categories and sampl $N$ images for each selected category to construct the training data. We experiment with different settings by adjusting the values of $M$ (\{4, 8, 12, 16\}) and N (\{2\}), observing how the model performes under these varying conditions. To gain a deeper understanding of the episode-based training method, we conduct a comparative analysis with the scenario of random sampling, where the mini-batch size is set to 32. The results in Table ~\ref{table3} indicate that the episode-based training method outperforms the random sampling training method, demonstrating superior performance. This suggests that our model achieves more effective generalization recognition for all categories by learning from the seen categories. Notably, when $M = 16$ and $N = 2$, the model attains the highest accuracy on AWA2 and CUB, When $M = 8$ and $N = 2$, the best results are obtained on SUN, highlighting the superiority of this training method.
\begin{table}[htbp]
  \centering
  \caption{Evaluation of the training method's influence on ZSL(ACC) and GZSL(H) results (\%). The most favorable outcomes are emphasized in \textbf{bold}.}
  \setlength{\tabcolsep}{1pt} 
    \begin{tabular}{c|cc|cc|cc|cc}
    \toprule
    \multirow{2}[4]{*}{Training Method} & \multirow{2}[4]{*}{M-way} & \multirow{2}[4]{*}{N-shot} & \multicolumn{2}{c|}{CUB} & \multicolumn{2}{c|}{AWA2} & \multicolumn{2}{c}{SUN} \\
\cmidrule{4-9}          &       &       & \textbf{ACC}   & \textbf{H}     & \textbf{ACC}   & \textbf{H}     & \textbf{ACC}   & \textbf{H} \\
    \midrule
    Random Sampling & /     & /     & 75.9  & 72.1    &72.4       &73.7       &64.3       &41.4  \\
    \midrule
    \multirow{4}[2]{*}{Episode-based} & 4     & 2     & 74.7  & 66.8  & 70.9  & 69.6  & 64.3  & 37.6 \\
          & 8     & 2     & 75.7  & 71.1  & 71.0    & 72.5  & \textbf{67.6} & \textbf{44.9} \\
          & 12    & 2     & 76.9  & 73.5  & 72.4  & 74.4  & 65.4  & 42.0 \\
          & 16    & 2     & \textbf{77.9} & \textbf{74.3} & \textbf{74.6} & \textbf{75.5} & 65.3  & 43.0 \\
    \bottomrule
    \end{tabular}%
  \label{table3}%
\end{table}%

\begin{figure}[h]
  \centering
  \subfigure{
  \begin{minipage}[t]{0.45\linewidth}
  \centering
  \includegraphics[width=1.0\linewidth]{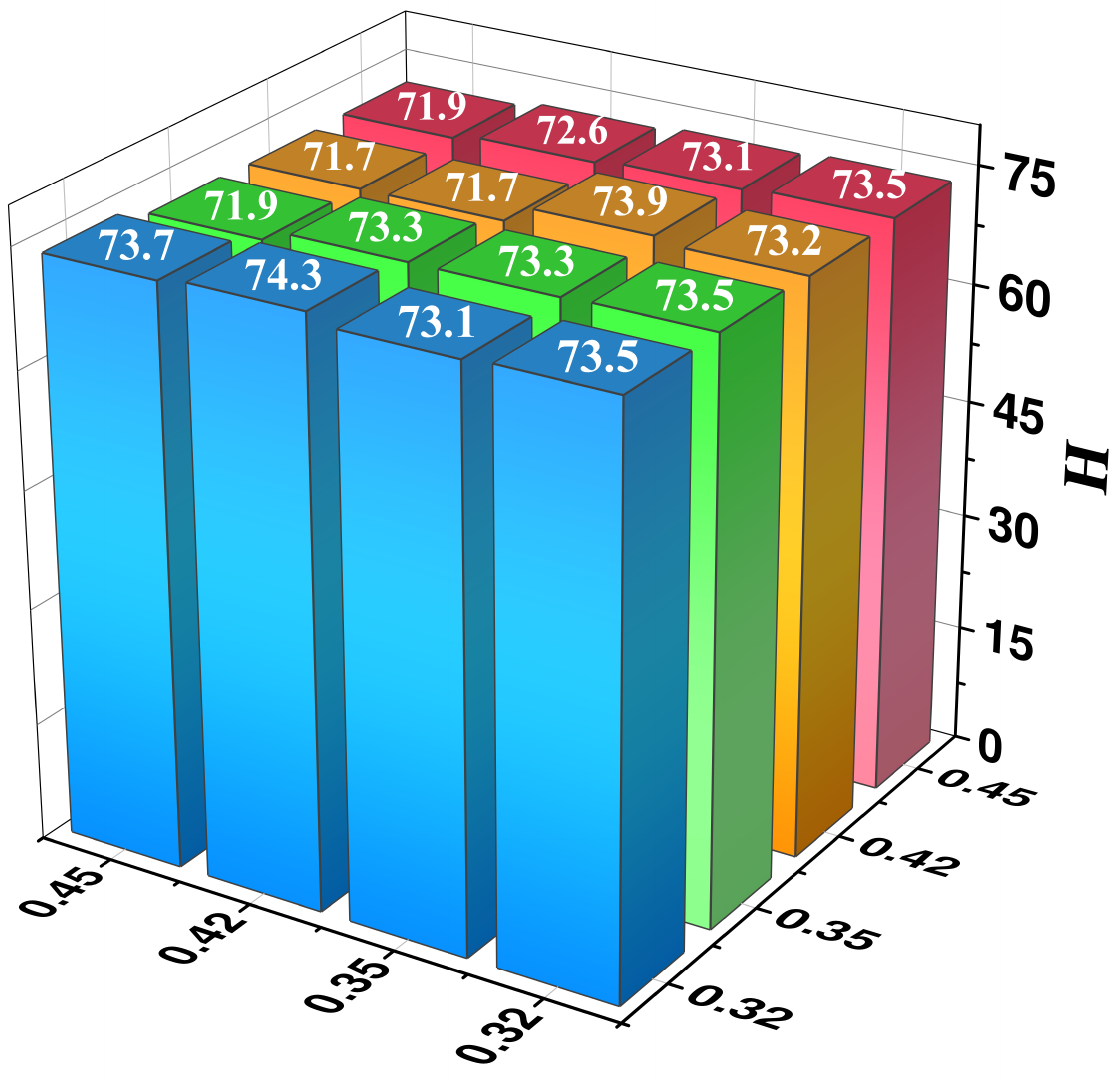}
  \end{minipage}
  }
  \subfigure{
  \begin{minipage}[t]{0.45\linewidth}
  \centering
  \includegraphics[width=1.0\linewidth]{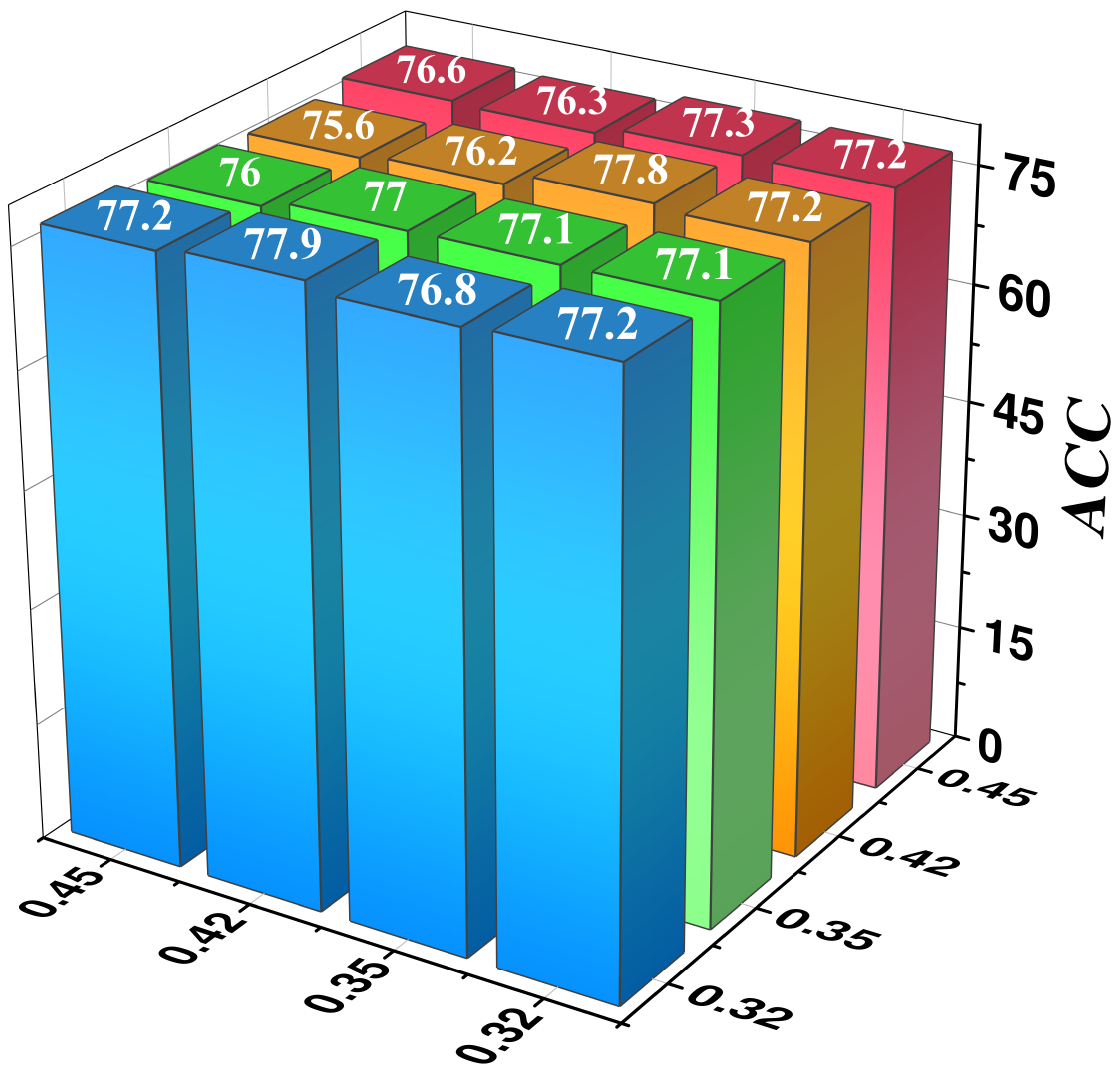}
  \end{minipage}
  }
  \centering
  \caption{ Evaluation Results for CUB with Different Values of $\mu$ and $\varepsilon$.}
  \label{fig4}
\end{figure}

\textbf{The Impact of Hard Samples Selection.} In the context of contrastive learning \cite{robinson2020contrastive}, the thoughtful selection of positive and negative samples is crucial for the success of HDAFL. Within the same class, samples demonstrating lower correlation are termed hard positives, whereas samples manifesting higher correlation across different classes are assigned the label of hard negatives. The careful selection of these challenging samples plays a key role in enhancing the discriminative power of the attribute embedding space. We conducted experiments on the CUB dataset for CZSL and GZSL, varying the values of $\mu$ and $\varepsilon$ in the range \{0.32,0.35,0.42,0.45\}, with all other parameters set to their default values. Figure ~\ref{fig4} illustrates that our model exhibits suboptimal performance when $\mu$ is set excessively large or $\varepsilon$ is set overly small. This is primarily due to a significant imbalance, where the number of positive samples is much smaller than that of negative samples. Weakening the attribute embedding space's robustness and discriminative potential occurs when there is an excessive removal of easy positives and an inadequate representation of easy negatives. It is evident from the experimental findings that our model achieves the highest performance in the GZSL task on the CUB dataset when $\mu$ is set to $0.32$ and $\varepsilon$ is set to $0.42$. To simplify, we opted for the same parameter values on SUN and AWA2.

\begin{figure}[h]
	\centering
	\includegraphics[width=1.0\linewidth]{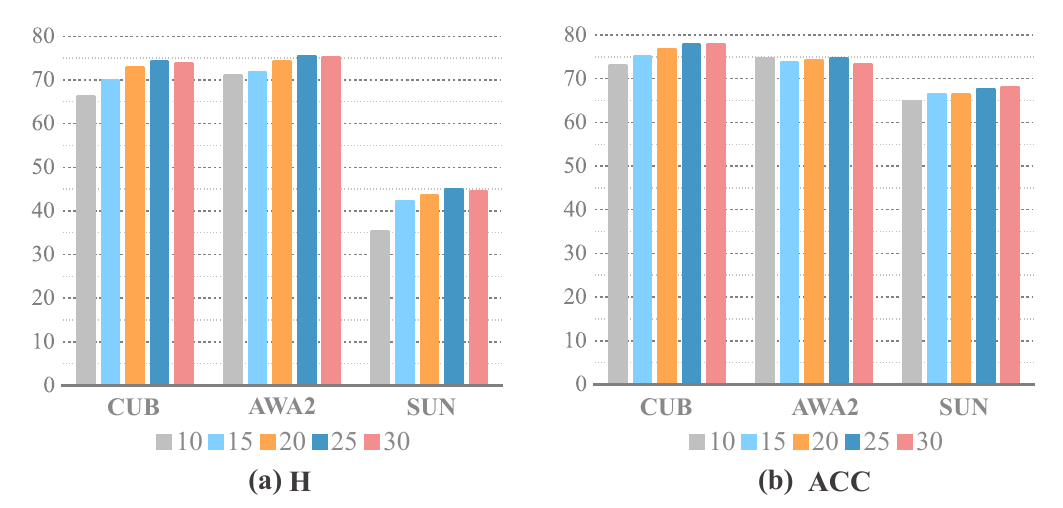}
	\caption{Analysis of the correlation between $\sigma$ values and the results ($\%$) in ZSL (ACC) and GZSL (H) across CUB, SUN, and AWA2 datasets.}\label{figsf}
\end{figure}

\textbf{The Impact of the Scaling Factor $\sigma$.} Figure ~\ref{figsf} illustrates the results of our model for $H$ and $ACC$ under different values of $\sigma$ (\{10, 15, 20, 25, 30\}) in the context of GZSL/ZSL. On three datasets, our method performs optimally when $\sigma$ is set to 25. Therefore, in our experiments, we fix $\sigma$ at 25 for all three datasets. This choice is based on a comprehensive analysis of the experimental results, ensuring optimal performance across different datasets.

\subsection{Qualitative Results}
\begin{figure}[h]
	\centering
	\includegraphics[width=0.5\textwidth]{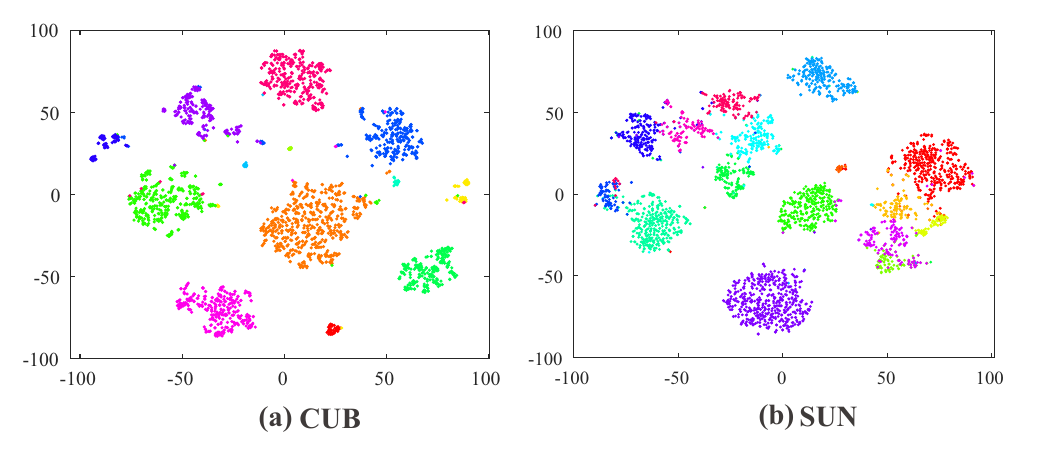}
	\caption{The visualization of attribute features using T-SNE on CUB and SUN datasets.}\label{fig7}
\end{figure}

\textbf{t-SNE Visualizations.}  To assess the efficacy of our HDAFL in distinguishing different attributes in the attribute embedding space, we opted for multiple images from the CUB and SUN datasets. Subsequently, $15$ attributes corresponding to each dataset were randomly chosen, and the features of these attributes were visualized using the t-SNE method \cite{van2008visualizing}. Upon observing Figure ~\ref{fig7}, it is evident that the features of different attributes exhibit distinct separation in the embedding space, indicating the model's capability to effectively differentiate them. Simultaneously, features of the same attribute are notably aggregated together. This observation strongly confirms that our model can capture common attribute features across images, enabling accurate attribute recognition and providing robust support for addressing domain shift issues.

\section{Conclusion}

In this study, we introduce HDAFL to boost the generalization capability of ZSL. HDAFL utilizes a contrastive learning mechanism to learn attribute and visual features. Further optimization of attribute features is achieved through a transformer-based attribute discrimination encoder. This ensures that the same attribute is brought together in the embedding space across different images, improving the distribution around corresponding attribute prototypes. This process enhances attribute discriminative power and addresses domain shift issues often overlooked in attention-based approaches. The localized discriminative attributes ultimately enhance the global representation of image features in ZSL. HDAFL demonstrates outstanding or competitive performance on three datasets, validating the effectiveness of the learned discriminative attributes.






\bibliographystyle{ACM-Reference-Format}
\bibliography{sample_base}

\end{document}